\documentclass[conference]{IEEEtran}
\IEEEoverridecommandlockouts
\usepackage{cite}
\usepackage{amsmath,amssymb,amsfonts}
\usepackage{algorithmic}
\usepackage{graphicx}
\usepackage{textcomp}
\usepackage[dvipsnames]{xcolor}
\usepackage{xcolor,colortbl}
\usepackage{multirow}
\usepackage{siunitx}
\usepackage{pgfplots}
\usepackage{subcaption}
\usepackage{pgfplotstable}
\usepackage{makecell}
\usepackage{hyperref}
\usepackage{tikz}
\usetikzlibrary{patterns}

\pgfplotsset{compat=1.3}
\usepgfplotslibrary{groupplots}

\def\BibTeX{{\rm B\kern-.05em{\sc i\kern-.025em b}\kern-.08em
    T\kern-.1667em\lower.7ex\hbox{E}\kern-.125emX}}

\begin{document}

\title{A Hybrid Learner for Simultaneous Localization and Mapping\\}

\author{Thangarajah~Akilan,~\IEEEmembership{Member,~IEEE}, Edna Johnson, Japneet Sandhu, Ritika Chadha, Gaurav Taluja%

% \thanks{T. Akilan, E. Johnson, G. Taluja, J. Sandhu, and R. Chadha are with the Department of Computer Science, Lakehead University, Thunder Bay, ON, Canada. (e-mail: \{takilan, ejohnso8, jsandhu6, rchadha1, gtaluja\}@lakeheadu.ca).}

}

\maketitle

\begin{abstract}

Simultaneous localization and mapping (SLAM) is used to predict the dynamic motion path of a moving platform based on the location coordinates and the precise mapping of the physical environment. SLAM has great potential in augmented reality (AR), autonomous vehicles, viz.  self-driving cars, drones, Autonomous navigation robots (ANR). This work introduces a hybrid learning model that explores beyond feature fusion and conducts a multimodal weight sewing strategy towards improving the performance of a baseline SLAM algorithm. It carries out weight enhancement of the front end feature extractor of the SLAM via mutation of different deep networks' top layers. At the same time, the trajectory predictions from independently trained models are amalgamated to refine the location detail. Thus, the integration of the aforesaid early and late fusion techniques under a hybrid learning framework minimizes the translation and rotation errors of the SLAM model. This study exploits some well-known deep learning (DL) architectures, including ResNet18, ResNet34, ResNet50, ResNet101, VGG16, VGG19, and AlexNet for experimental analysis. An extensive experimental analysis proves that hybrid learner (HL) achieves significantly better results than the unimodal approaches and multimodal approaches with early or late fusion strategies. Hence, it is found that the Apolloscape dataset taken in this work has never been used in the literature under SLAM with fusion techniques, which makes this work unique and insightful. \\

\end{abstract}


\section{Introduction}\label{intro}
SLAM is a technological process that enables a device to build a map of the environment, at the same time, helps compute the relative location on predefined map. 
% It assists us in simultaneous position and corresponding mapping based on the already trained input to the system. 
It can be used for range of applications from self-driving vehicles (SDV) to space and maritime exploration, and from indoor positioning to search and rescue operations. The primary responsibility of a SLAM algorithm is to produce an understanding of a moving platform's environment and the location of the vehicle by providing the value of its coordinates; thus, improving the formation of a trajectory to determine the view at a particular instance. As a SLAM is one of the emerging technologies, numerous implementations have been introduced but the DL-based approaches surmount others by their efficiency in extracting the finest features and giving better results even in a feature-scarce environment. 
% composed of distinguished landmarks.

This study aims to improve the performance of a self-localization module based on PoseNet~\cite{kendall2015posenet} architecture through the concept of hybrid learning that does a multimodal weight mutation for enhancing the weights of a feature extractor layer and refines the trajectory predictions using amalgamation of multimodal scores. The ablation study is carried out on the Apolloscape~\cite{wang2018dels, wang2019apolloscape}, as per our knowledge, there has been no research work performed on the self-localization repository of the Apolloscape dataset, in which the proposed HL has been evaluated extensively. 
% .  extraction from the feature extractor models.
% on Apolloscape dataset. As per our knowledge, there has been no research work performed on the self-localization module of the Apolloscape dataset which defines its performance analysis at its best.  
% To obtain the best performance results a fusion technique is exercised into consideration for our study. 
% There have been numerous research works done around feature fusion. Most of the handcrafted feature fusion strategies use mathematical operations to integrate various features.
% Our study practices the concept of weight extraction from the feature extractor models. 
The experimental analysis presented in this work consists of three parts, in which initial two parts form the base for the third. The first part concentrates on an extensive evaluation of several DL models, as feature extractors. The second part analyzes of two proposed multimodal fusion approaches: (i). an early fusion via layer weight enhancement of feature extractor, and (ii). a late fusion via score refinement of the trajectory (pose) regressor. Finally, the third part aims at the combination of early and late fusion models forming a hybrid learner with addition or multiplication operation. Here, the late fusion model harnesses five pretrained deep convolutional neural networks (DCNNs), viz. ResNet18, ResNet34, ResNet101. VGG16, and VGG19 as the feature extractor for pose regressor module. While, the early fusion model and the HL focuses on exploiting the best DCNNs, ResNet101 and VGG19 based on their individual performance on the Apolloscape self-localization dataset.

When analyzing the results of the early and late fusion models, it is observed that the early fusion encompasses $14.842 m$ of translation error and $0.673^\circ$ of rotation error. On the other hand, the late fusion achieves $9.763 m$ of translation error and $0.945^\circ$ of rotation error. On analyzing the hybrid learners, the additive hybrid learner (AHL) gets $10.400 m$ of translation error and $0.828^\circ$ of rotation error, whereas the multiplicative hybrid learner (MHL) records $9.307 m$ and $1.206^\circ$ of translation and rotation errors, respectively. By fusing the predictions of AHL and MHL called hybrid learner full-fusion (HLFF) produces better results than all other models with $7.762 m$ and $0.886^\circ$ of translation rotation errors, respectively. 
% The average is determined by considering the results of hybrid fusion model of addition on ResNet101 and VGG19 and multiplication on ResNet101 and VGG19.

The rest of the paper is organized as follows. Section~\ref{literature} reviews relevant SLAM literature and provides basic detail of the PoseNet, unimodality, and multimodality. Section~\ref{proposed} elaborates the proposed hybrid learner including required pre-processing operations. Section~\ref{results} describes the experimental setup and analyzes the obtained results from various models. Section~\ref{conclusion} concludes the research work.

\section{Background}\label{literature}
\subsection{SLAM}

Simultaneous localization and mapping is an active research domain in robotics and artificial intelligence (AI). 
% As discussed in the section background of chapter I, SLAM
It enables a remotely automated moving vehicle to be placed in an unknown environment and location. According to Whyte~\textit{et al.}~\cite{Durrant-Whyte06simultaneouslocalisation} and Montemerlo~\textit{et al.}~\cite{montemerlo2003fastslam}, SLAM should build a consistent map of this unknown environment and determine the location relative to the map. Through SLAM, robots and vehicles can be truly and completely automated without any or minimal human intervention. But the estimation of maps consists of various other entities, such as large storage issues, precise location coordinates, which makes SLAM a rather intriguing task, especially in the real-time domain.

Many researches have been done worldwide to determine the efficient method to perform SLAM. In~\cite{montemerlo2002fastslam}, Montemerlo  {\textit{et al.}} propose a model named FastSLAM, as an efficient solution to the problem. FastSLAM is a recursive algorithm that calculates the posterior distribution spanning over autonomous vehicle’s pose and landmark locations, yet, it scales logarithmically with the total number of landmarks. This algorithm relies on an exact factorization of the posterior into a product of landmark distributions and a distribution over the paths of the robot. The research on SLAM originates on the work of Smith and Cheeseman~\cite{smith1986representation} that propose the use of the extended Kalman filter (EKF). It is based on the notion that pose errors and errors in the map are correlated, and the covariance matrix obtained by the EKF represents this covariance. There are two main approaches for the localization of an autonomous vehicle: metric SLAM and appearance-based SLAM~\cite{kendall2015posenet}. However, this research focuses on the appearance-based SLAM that is trained by giving a set of visual samples collected at multiple discrete locations.

\subsection{PoseNet}

% \begin{table}[t]
%     \centering
%     \setlength{\tabcolsep}{3pt}

%     \begin{tabular}{|c|ccc|c|}
%     \hline
%     \cellcolor{green!50}\textbf{Front-end} & \multicolumn{4}{c|}{ \cellcolor{red!40}\textbf{Back-end}} \\
    
%     \hline
%     \cellcolor{green!20}{Feature Extractor:}  & \multicolumn{3}{c|}{\cellcolor{red!30}Pose Regressor} & \cellcolor{red!25}Output \\
%     \cellcolor{green!20}A pretrained CNN & \multicolumn{3}{c|}{\cellcolor{red!10}Dropout$\rightarrow$ Pooling$\rightarrow$ Dense} & \cellcolor{red!10} Translation \& Rotation \\
%     \hline
% \end{tabular}
%     \caption{PoseNet Architecture.}
%     \label{fig:posenet_archi}
% \end{table}

The neural network (NN) comprises of several interconnected nodes and associated parameters, like weights and biases. The weights are adjusted through a series of trials and experiments in the training phase so that the network can learn and can be used to predict the outcomes at a later stage.
% for the real-time test data. 
There are various kinds of NN's available, for instance, Feed-forward neural network (FFNN), Radial basis neural network (RBNN), DCNN, Recurrent neural network (RNN), etc. 
% FeedForward Neural Network is the deep learning models which aim to approximate a function. They are called as feedforward because information flows from input through the intermediate computations to the final output. It does not involve any feedback connections in which the outputs of the successive layer are fed to the preceding layer~\cite{Goodfellow-et-al-2016}. But Martin {\textit{et al.}}~\cite{Sundermeyer_lstmneural} proposes an illustration where Long Short Term Memory (LSTM) Neural Networks are more efficient than feedforward Neural Network in the domain of Language Modelling. 
Among them, the DCNN's have been highly regarded for the adaptability and finer interpretability with accurate and justifiable predictions in applications range from finance to medical analysis and from science to engineering. Thus, the PoseNet model for SLAM shown in Fig.~\ref{fig:posenet_archi}, harness the DCNN to be firm against difficult lighting, blurredness, and varying camera instincts~\cite{kendall2015posenet}.
% , which overcomes all the above-discussed issues by taking an image as the input and regressing the poses of the image taken. 
% Furthermore, it is observed that PoseNet is a robust model that localizes from high-level features and is firm against difficult lighting, blurredness, and different camera instincts~\cite{kendall2015posenet}. The block diagram is shown in 
Figure~\ref{fig:posenet_archi} depicts the underlying architecture of the PoseNet. It subsumes a front-end with a feature extractor and a back-end with a regression subnetwork. The feature extractor can be a pretrained DCNN, like ResNet$34$, VGG$16$, or AlexNet. The regression subnetwork consists of three stages: a dropout, an average pooling, a dense layer interconnected, sequentially. It receives the high dimensional vector from the feature extractor. Through the average pooling and dropout layers, it is then reduced to a lower dimension for generalization and faster computation \cite{DBLP:journals/corr/WalchHLSHC16}.
The predicted poses are in Six-degree of freedom (6-DoF), which define the six parameters in translation and rotation~\cite{kendall2015posenet}. The translation consists of forward-backward, left-right, and up-down parameters forming the axis of 3D space as $x-axis$, $y-axis$, and $z-axis$, respectively. Likewise, the rotation includes yaw, pitch, and roll parameters of the same 3D space noted as $normal-axis$, $transverse-axis$, and $longitudinal-axis$, respectively. 
% These core parameters help in getting the accurate position of the vehicle and then form the trajectory.

\begin{figure}[t]
    \centering
    % \includegraphics[trim={0.0cm, 0.0cm, 0.0cm, 0.0cm}, clip, width=1.0\columnwidth]{posenet.png}
    % \vspace{-0.5cm}
    % \setlength{\tabcolsep}{3pt}
    \resizebox{1\columnwidth}{!}{%
    \begin{tabular}{|ccccc|}
    \hline
    \cellcolor{green!50}\textbf{Front-end} & \multicolumn{4}{c|}{ \cellcolor{red!40}\textbf{Back-end}} \\
    
    \hline
    \cellcolor{green!20}{\textbf{\textcolor{ForestGreen}}{Feature Extractor:}}  & \multicolumn{3}{c|}{\cellcolor{red!30}\textbf{\textcolor{BrickRed}{Pose Regressor}}} & \cellcolor{red!25}\textbf{\textcolor{BrickRed}{Output: Poses}} \\
    \cellcolor{green!20}A pretrained CNN & \multicolumn{3}{c|}{\cellcolor{red!10}Dropout$\rightarrow$ Pooling$\rightarrow$ Dense} & \cellcolor{red!10} Translation \& Rotation \\
    \hline
    \end{tabular}% <------ Don't forget this %
}
    \caption{PoseNet Architecture Subsuming a Feature Extractor and a Pose Regressor Subnetwork.}
    
    \label{fig:posenet_archi}
    % \vspace{-0.5cm}
\end{figure}

Then, these six core parameters are converted to seven coordinates: $x_1$, $x_2$, and $x_3$ of translation coordinates, and $y_1, y_2, y_3, y_4$ of rotation coordinates. It is because the actual rotation poses are in Euler angles. Thus, a pre-processing operation converts the Euler angles into quaternions. The quaternions are the set of four values ($x_o$, $y_1$, $y_2$ and $y_3$), where $x_o$ represents a scalar rotation of the vector - $y_1$, $y_2$ and $y_3$. This conversion is governed by the expressions given in Eq.~(\ref{eq:10}) - (\ref{eq:13}).  
% Function mat2quat from transforms3d.quaternions package performs the quaternion conversion, which takes parameters and returns the output as following:
% \subsubsection{Parameters} array-like 3×3 rotation matrix.
% \subsubsection{Returns} (4,) array closest quaternion to input matrix.

% Function mat$2$quat performs the conversion of euler angles to quaternions as given in the equations below:
\begin{equation} \label{eq:10}
x_0 = (\sqrt{1 + c_1c_2 + c_1c_3 - s_1s_2s_3 + c_2c_3})/2,
\end{equation} 
\begin{equation} \label{eq:11}
y_1 = (c_2s_3 + c_1s_3 + s_1s_2c_3)/4x_0,
\end{equation} 
\begin{equation} \label{eq:12}
y_2 = (s_1c_2 + s_1c_3 + c_1s_2s_3)/4x_0,
\end{equation} 
\begin{equation} \label{eq:13}
y_3 = (-s_1s_3 + c_1s_2c_3 + s_2)/4x_0,
\end{equation} 
where $c_1$ = $\cos(roll/2)$, $c_2$ = $\cos(yaw/2)$,  $c_3$ = $\cos(pitch/2)$,  $s_1$ = $\sin(roll/2)$,  $s_2$ = $\sin(yaw/2)$, and  $s_3$ = $\sin(pitch/2)$.

% \textcolor{red}{As an output, 6 DoF regressor with camera translation ($x_1, x_2, x_3$) and camera rotation ($y_1, y_2, y_3, y_4$) is obtained.}

% \textcolor{red}{The fully connected layer in the PoseNet architecture yields high dimensionality output which is carefully dealt in PoseNet through dropout layers \cite{cnnpluslstm}}.

% \subsection{Loss Function}

The pose regressor subnetwork is to be trained to minimize the translation and rotation errors. These errors are combined into a single objective function, $L_\beta$ as defined in Eq.~(\ref{eq:1})~\cite{kendall2017geometric}.
\begin{equation} \label{eq:1}
L_\beta(I) = L_x(I) + \beta L_q(I),
\end{equation} 
where $L_x$, $L_q$ are the losses of translation and rotation respectively, and $I$ is the input vector representing the discrete location in the map. $\beta$ is a scaling factor that is used to balance both the losses and calculated using homoscedastic uncertainty that combines the losses as defined in~(\ref{eq:2}). 
\begin{equation} \label{eq:2}
L_\sigma(I) = \frac{L_x(I)}{\hat\sigma_x^2} + \log\hat\sigma_x^2 + \frac{L_q(I)}{\hat\sigma_q^2} + \log\hat\sigma_q^2,
\end{equation}
where $\hat{\sigma}_x$ and $\hat{\sigma}_q$ are the uncertainties for translation and rotation respectively. Here, the regularizers  $\log\hat\sigma_x^2$ and $\log\hat\sigma_q^2$ prevent the values from becoming too big \cite{kendall2017geometric}. It can be calculated using a more stable form as in Eq.~(\ref{eq:3}), which is very handy for training the PoseNet.
\begin{equation} \label{eq:3}
L_\sigma(I) = L_x(I)^{ -\hat{s}_x} + \hat{s}_x  + L_q(I)^{ -\hat{s}_q} + \hat{s}_q,
\end{equation}
where the learning parameter $s=\log{\hat\sigma^2}$. Following \cite{kendall2017geometric}, in this work, $\hat{s_x}$ and $\hat{s_q}$ are set to $0$ and $- 3.0$, respectively. 
% to get the best results .

\subsection{The Front-end Feature Extractor}

As discussed earlier the PoseNet take advantage of transfer learning (TL), whereby it uses pretrained DCNN as feature extractor. 
TL differs from traditional learning, as, in latter, the models or tasks are isolated and function separately. They do not retain any knowledge, whereas TL learns from the older problem and leverages the new set of problems~\cite{taylor2009transfer}. Thus, in this work, versions of ResNet, versions of VGG, and AlexNet are investigated. Some basic information of these DCNN's are given in the following subsections.  
% Matthew \textit{et al.}~\cite{taylor2009transfer} gives a detailed survey about various transfer learning methods in Reinforcement Learning (RL) paradigm. 
% It surveys and groups transfer learning methods on the basis of the goal of the transfer method, metrics to measure the success, and what information should be transferred. It discusses various metrics to analyze the performance of the transfer learning method, such as jump-start, asymptotic performance, total reward, time to the threshold, and transfer ratio.

% Wenyuan et al. in~\cite{dai2007boosting} present a novel transfer learning algorithm called TrAdaBoost, to extend a boosting-based learning algorithm. It is handy in the domains when a new problem from a new dataset arises, but there exists a labeled data from the similar old dataset. It comes with two issues, firstly, labeling the new data would be too costly, and no proper use of old data would lead to a waste of data. Now, this new algorithm proposed by authors in this paper, would take a small amount of new data and utilizing a big chunk of old data to construct a high-quality classification model. In this way, TrAdaBoost effectively transfers knowledge from old data to new data.

\subsubsection{AlexNet}
It was the winner in 2012 ImageNet Large Scale Visual Recognition Competition (ILSVRC'12) with a breakthrough performance \cite{AlexNet}. It consists of five convolution (Conv) layers taking up to 60 million trainable parameters and 650,000 neurons making it one of the huge models in DCNN's. The first and second Conv layers are followed by a max pooling operation. But the third, fourth, and fifth Conv layers are connected directly, and the final stage is a dense layer and a thousand-way Softmax layer. It was the first time for a DCNN to adopt rectified linear units (ReLU) instead of the tanh activation function and to use of dropout layer to eradicate overfitting issues of DL.

\subsubsection{VGG (16, 19)}
Simonayan and Zisserman~\cite{VGG} proposed the first version of VGG network named VGG16 for the ILSVRC'14. It stood 2nd in the image classification challenge with the error of top-5 as $7.32\%$. VGG16 and 19 consist of 16 and 19 Conv layers, respectively with max pooling layer after set of two or three Conv layers. It comprises of two fully connected layers and a thousand-way Softmax top layer, similar to AlexNet. The main drawbacks of VGG models are high training time and high network weights. 

\subsubsection{ResNet (18, 34, 50, 101)}
ResNet18~\cite{resNet18} was introduced to compete in ILSVRC'15, where it outperformed other models, like VGG, GoogLeNet, and Inception. All the ResNet models used in this work are trained on the ImageNet database that consists more than million images. Experiments have depicted that even though ResNet18 is a subspace of ResNet34, yet its performance is more or less equivalent to ResNet34. ResNet18, 34, 50, and 101 consist of 18, 34, 50, and 101 layers, respectively. This paper, firstly, evaluates the performance of the PoseNet individually using the above mentioned ResNet models besides other feature extractors. Consequently, it chooses the best ones to be used in the fusion modalities and in the hybrid learner, thereby, establishing a good trade-off between depth and performance. The ResNet models constitute of residual blocks, whereas ResNet18 and 34 have two stack of deep residual blocks, while ResNet50 and 101 have three deep residual blocks. A residual block subsumes five convolutional stages, which is followed by average pooling layer. Hence, each ResNet model has a fully connected layer followed by a thousand-way Softmax layer to generate a thousand-class labels.

\subsection {Multimodal Feature Fusion}

%Mutimodality fusion is one of the effective and widely adopted techniques used across various learning systems to improve the performance~\cite{akismc, akiccece}. Whereby, different models or the complementary cues from the models are combined using strategies, such as mathematical modeling and subspace transformation~\cite{xufusion}. %
There are many existing researches that have taken the advantage of various strategies for feature extraction and fusion. For an instance, Xu~\textit{et al.}~\cite{firstLS} modify the Inception-ResNet-v1 model to have four layers followed by a fully connected layer in order to reduce effect of overfitting, as their problem domain has less number of samples and fifteen classes. 
% the output to 64. The last layer i.e. the softmax layer generates the class labels only for the 15 classes
% to extract the features. But since the dataset used for the work was too small and just comprised of 15 classes, therefore, Inception-ResNet-v1 was modified to alleviate the overfitting. Six convolutional layers in the model is reduced down to. 
On the other hand, Akilan~\textit{et al.}~\cite{secondLS} continue with a TL technique in the feature fusion, whereby they extract features using multiple DCNN's, namely AlexNet, VGG16 and Inception-v3. 
As these extractors will result into a varied feature dimensions and sub-spaces, feature space transformation and energy-level normalisation are performed to embed the features into a common sub-space using dimensionality reduction techniques like PCA. Finally, the features are fused together using fusion rules, such as concatenation, feature product, summation, mean value pooling, and maximum value pooling. 

Fu~{\textit{et al.}}~\cite{thirdLS} also consider the dimension normalization techniques to produce a consistently uniform dimensional feature space. It presents supervised and unsupervised learning sub-space learning method for dimensionality reduction and multimodal feature fusion. The work also introduces a new technique called, Tensor-Based discriminative sub-space learning. This technique gives better results, as it produces the final fused feature vector of adequate length, i.e., the long vector if the number of features are too large and the shorter vector if number of features are small. Hence, Bahrampour~{\textit{et al.}}~\cite{fourthLS} introduce a multimodal task-driven dictionary learning algorithm for information that is obtained either homogenously or heterogeneously. These multimodal task-driven dictionaries produce the features from the input data for classification problems.

\subsection{Hybrid Learning}

Sun~\textit{et al.}~\cite{sun2013hybrid} proposes a hybrid convolutional neural network for face verification in wild conditions. 
% In this paper, a new approach is proposed wherein, feature extraction and recognition are combined in one architecture. 
Instead of extracting the features separately from the images, the features from the two images are jointly extracted by filter pairs. The extracted features are then processed through multiple layers of the DCNN to extract high-level and global features. The higher layers in the DCNN discussed in their work locally share the weights, which is quite contrary to conventional CNNs. In this way, feature extraction and recognition are combined under the hybrid model.  

Similarly, Pawar~\textit{et al.}~\cite{pawar2010hybrid} develope an efficient hybrid approach involving invariant scale features for object recognition. In the feature extraction phase, the invariant features, like color, shape, and texture are extracted and subsequently fused together to improve the recognition performance. The fused feature set is then fed to the pattern recognition algorithms, such as support vector machine (SVM), discriminant canonical correlation, and locality preserving projections, which likely produces either three distinct or identical numbered false positives. To hybridize the process entirely, a decision module is developed using NN's that takes in the match values from the chosen pattern recognition algorithm as input, and then returns the result based on those match values. 

\begin{figure*}[t]
    \centering
    \includegraphics[trim={0.0cm, 0.5cm, 0.0cm, 0.0cm}, clip, width=\textwidth]{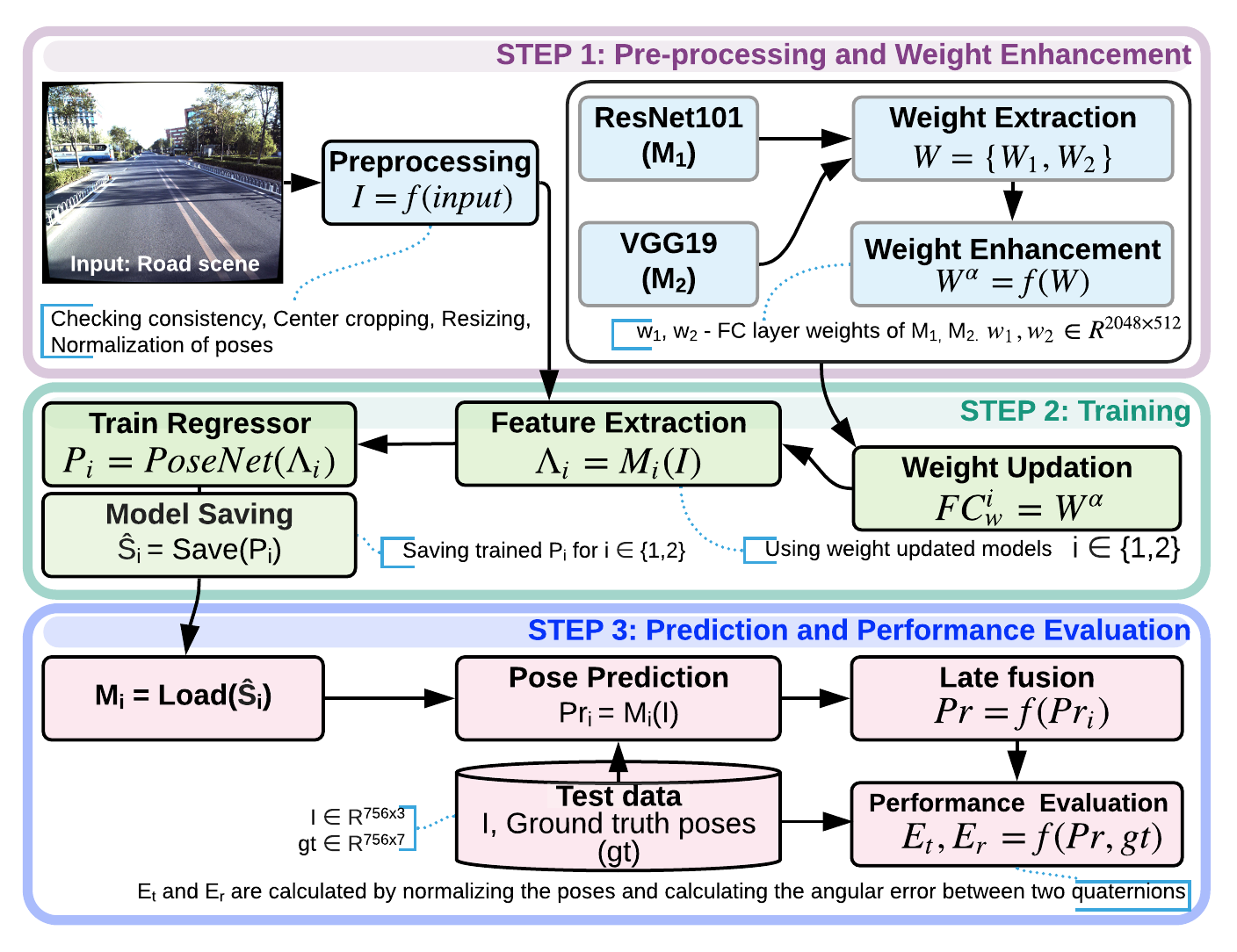}
    %\vspace{-0.3cm}
    \caption{Operational Flow of the Proposed Hybrid Learner with a Weight Sewing Strategy and a Late Fusion Phase of the Predicted Poses Towards Improving the Localization Capability of a SLAM Model - PoseNet.}
    \vspace{-0.3cm}
    \label{fig:hybrid}
\end{figure*}

However, the hybrid learner (Fig.~\ref{fig:hybrid}) introduced in this work is more unique and insightful than the existing hybrid fusion approaches. It focuses on enhancing and updating the weights of the pretrained unimodals before using them as front-end feature extractors of the PoseNet. Besides that it not only does a mutation of multimodal weights of the feature extraction dense layer, but also fuses the predicted scores of the pose regressor.

% (ResNet$101$ and VGG$19$) before training the model, which is followed by fusing the predicted scores obtained in the testing phase. Thus, hybrid fusion is the summation of early and late fusion under a single network architecture.

% Average score fusion is used in the score fusion model as explained in the subsection score fusion under proposed method. Once all the saved five models are loaded, the scores are averaged together to give the fused model.

\section{PROPOSED METHOD}\label{proposed}

% \subsection{PoseNet Architecture}
% The block diagram shown in Fig.~\ref{fig:posenet_archi} depicts the basic architecture of the PoseNet. It subsumes a pretrained DCNN as feature extractor and a regression subnetwork. The feature extractor can be a DCNN like ResNet, VGG or AlexNet. The regression subnetwork consists of  an average pooling, dropout, a fully connected layers networked sequentially. It receives the features extracted from the feature extractor and regresses the  poses. 

% As an output, 6 DoF regressor with camera translation ($x_1, x_2, x_3$) and camera rotation ($y_1, y_2, y_3, y_4$) is obtained.

% \begin{figure}[!ht]
%     \centering
%     %\includegraphics [trim=left bottom right top, clip]
%     \includegraphics[trim={2.5cm, 2.5cm, 2.5cm, 3.0cm}, clip, width=1.0\columnwidth]{posenet.png}
%     \caption{PoseNet Architecture.}
%     \label{fig:posenet_archi}
%     \vspace{-0.5cm}
% \end{figure}

\subsection{Hybrid Weight Swing and Score Fusion Model}
The Fig.~\ref{fig:hybrid} shows a detailed flow diagram of the hybrid learner. It consists of two parts, wherein the first part (Step 1 - weight enhancement) carries out an early fusion by layer weight enhancement of the feature extractor and the second part (Step 3) does a late fusion via score refinement of the models involved in the early fusion. The two best feature extractors chosen for early fusion based on their individual performances are used for forming the hybrid learner. The early fusion models obtained through fusing the dense layer weights of ResNet101 and VGG19 by addition or multiplication. In late fusion, the predicted scores of multiple pose regressors with the weight enhanced above feature extractors are amalgamated using average filtering to achieve better results.

% \subsection{Operational Flow}
% \subsubsection{Preprocessing}

\subsection{Preprocessing}

Before passing the images and poses to the PoseNet model, it is required to preprocess the data adequately. The preprocessing involves checking the consistency of the images, resizing and center cropping of the images, extraction of mean and standard deviation, and normalization of the poses.
% and filtering the images based on the training and validation. 
The images are resized to $260\times260$, and center cropped to $250\times250$. The translation is used to get the minimum, maximum, mean and standard deviation. The rotation values are read as Euler angles which suffer from wrap around infinities, gimbal lock and interpolation problems. To overcome these challenges, Euler rotations is converted to quaternions \cite{bouthellierrotations}.

% \subsection{Summary}
% The proposed solution introduces two strategies to enhance the performance of the PoseNet: Training time weight enhancement of \textcolor{red}{the feature extractor}, and late fusion of the predictions.  

\subsection{Multimodal Weight Sewing via Early Fusion (EF)}

% The camera feed consists of poses and images, which are center cropped, resized followed by the normalization of the poses. 
The preprocessed data is fed to the feature extractors: ResNet101 and VGG19. These two models are selected based on their individual performance on the Apolloscape test dataset. Using these two feature extractors the PoseNet has produced minimum translation and rotation errors, as recorded in Table~\ref{tab:table1}. The weights of the top feature extracting dense layers of the two feature extractors are fused via addition or multiplication operation. The fused values are used to update the weights of the respective dense layer of the ResNet101 and VGG19 feature extractors. The updated models are then used as new feature extractors for the regressor subnetwork. Then, the regressor is trained on the training dataset. 

% The validation dataset is later used for testing the model. The predicted results are compared with the ground truth results to calculate the translation and rotation losses. The normalization of the poses gives the translation error and the difference of quaternion angles gives the rotation error.

\subsection{Pose Refinement via Late Fusion (LF)}

The trained models with the updated ResNet101 and VGG19 using early fusion of multiplication and addition operations are moved onto the late fusion phase as shown in Step 3 in Fig.~\ref{fig:hybrid}. Where, the loaded weight enhanced early fusion models, simultaneously predict the poses for each input visual. The predicted scores from these models (in this case, ResNet101 and VGG19) are amalgamated with average filtering. This way of fusion is denoted as AHL. Similarly, the predicted scores of the early fusion models can be refined using multiplication and it is denoted as MHL. Finally, the predicted scores of the four early fusion models using addition and multiplication are fused together using average mathematical operation to achieve the predicted scores for the full hybrid fusion model stated earlier (Section~\ref{intro}) in the paper as HLFF. These predicted poses are then compared with the ground truth poses to calculate the mean and median of translation and rotational errors.

\section{Experimental Setup and Results}\label{results}

\subsection{Dataset}
Apolloscape dataset is used for many computer vision tasks related to autonomous driving. The Apolloscape dataset consists of modules including instance segmentation, scene parsing, lanemark parsing and self-localization that are used to understand and make a vehicle to act and reason accordingly~\cite{wang2019apolloscape, wang2018dels}. 
% In this research, we have used the Apolloscape dataset for self-localization to know the exact position of the vehicle and the trajectory traversed by it.
% such as semantic scene parsing, semantic lane mark segmentation, and self-localization \cite{wang2019apolloscape}. 
The Apolloscape dataset for self-localization is made up of images and poses on different roads at discrete locations. The images and the poses are created from the recordings of the videos.
% used while making the dataset. 
Each record of the dataset has multiple images and poses corresponding to every image. The road chosen for the ablation study of this research is \textit{zpark}. It consists of a total of 3000 stereo vision road scenes. For each image, there is a ground truth of poses with 6-DoF. 
From this entire dataset, a mutually exclusive training and test sets are created with the ratio of $3:1$.

% Two camera images clicked at the same timestamp from different angles are used. The camera poses are the 6-DoF poses which defines the three parameters of translation and rotation each.

\subsection{Evaluation Metric}

Measuring the performance of the machine learning model is pivotal to comparing the various CNN models. Since every CNN model is trained and tested on different datasets with varied hyperparameters, it is necessary to choose the right evaluation metric. 
% Choosing the right evaluation metrics depends on the type of model as well, i.e., if it is a regression model or a classification model. Alexei Botchkarev in~\cite{botchkarev2018performance} provides an overview of various performance metrics and their contribution to choosing the right CNN model. It divides metrics under primary metrics, extended metrics, composite metrics, and hybrid metrics. Primary metrics comprise of Mean Absolute Error (MAE), Mean Squared Error (MSE), etc. Whereas extended metrics consist of Normalized Root Mean Squared Error (NRMSE), and composite metrics consist of Mean Absolute Scaled Error (MASE), Relative Mean Absolute Scaled Error (RMAE). Hybrid metrics means when two or more performance metrics are used simultaneously in the same experiment.
As the domain of this work is a regression problem, the mean absolute error (MAE) is used to measure the performance of the set of models ranging from unimodals to the proposed hybrid learner. MAE is a linear score, which is calculated as an average of the absolute difference between the target variables and the predicted variables using the formula given in Eq.~(\ref{eq:MAE}).
\begin{equation}\label{eq:MAE}
    MAE = \frac{1}{n}\sum_{i=1}^{n}\mid{x_i - x}\mid,
\end{equation}
where, $n$ is the total number of samples in the validation dataset, $x_i$ and $x$ are the predicted and ground truth poses, respectively. Since, it is an average measure, it implies that all the distinct values are weighted equally ignoring any bias involvement. 

\begin{table}[t]
 \begin{center}
 \setlength\tabcolsep{4pt}
%  {\setlength{\extrarowheight}{4pt}%
 \begin{tabular}{|l|c|c|c|c|c|c|}
 \hline
 \makecell{\textbf{Model Name}} & \makecell{\textbf{Median} \\$\mathbf{e_t (m)}$} & \makecell{\textbf{Mean} \\$\mathbf{e_t (m)}$} & \makecell{\textbf{Median} \\ $\mathbf{e_r (^\circ)}$} & \makecell{\textbf{Mean} \\$\mathbf{e_r (^\circ)}$} & \makecell{\textbf{MAPST} \\$\mathbf{(s)}$} \\

\hline \hline
M$1$ - ResNet$18$ & $21.194$ & $24.029$ & $0.778$ & $0.900$ & $0.092$ \\
\hline
M$2$ - ResNet$34$ & $20.990$ & $23.597$ & $0.673$ & $0.824$ & $0.093$ \\
\hline
M$3$ - ResNet$50$ & $18.583$ & $20.803$ & $0.903$ & $1.434$ & $0.095$ \\
\hline
M$4$ - ResNet$101$ & $16.227$ & $19.427$ & $0.966$ & $1.230$ & $0.098$ \\
\hline
M$5$ - VGG$16$ & $17.150$ & $21.571$ & $1.079$ & $1.758$ & $0.103$ \\
\hline
M$6$ - VGG$19$ & $16.820$ & $19.935$ & $0.899$ & $1.378$ & $0.111$ \\
\hline
M$7$ - AlexNet & $46.992$ & $53.004$ & $4.282$ & $7.177$ & $0.108$ \\
\hline
M$8$ - LF & $9.763$ & $10.561$ & $0.945$ & $4.645$ & $0.146$ \\
\hline
M$9$ - AEF$^{ResNet101}$  & $14.870$ & $18.256$ & $0.673$ & $0.784$ & $0.134$ \\
\hline
M$10$ - MEF$^{ResNet101}$ & $14.842$ & $18.013$ & $0.779$ & $0.977$ & $0.131$  \\
\hline
M$11$ - AEF$^{VGG19}$ & $11.047$ & $13.840$ & $0.742$ & $1.024$ & $0.137$ \\
\hline
M$12$ - MEF$^{VGG19}$ & $10.730$ & $14.181$ & $0.756$ & $1.141$ & $0.135$ \\
\hline
M$13$ - AHL & $10.400$ & $12.193$ & $0.828$ & $5.155$ & $0.142$ \\
\hline
M$14$ - MHL & $9.307$ & $11.420$ & $1.206$ & $5.455$ & $0.142$ \\
\hline
M$15$ - HLFF & $7.762$ & $8.829$ & $1.008$ & $4.618$ & $0.144$ \\
\hline \hline
\end{tabular}%}
\end{center}
\caption{Performance Analysis of Various Models: $e_t$ - translation error, $e_r$ - rotation error, MAPST - mean average per sample processing  time.}
%\vspace{-0.3cm}
\label{tab:table1}
\end{table}

\subsection{Performance Analysis}

This Section elaborates the results obtained from each of the model introduced earlier in this paper. 
% The Apolloscape dataset is run through different DCNNs, which include ResNet18, ResNet34, ResNet50, ResNet101, VGG16, VGG19, and AlexNet. 

\subsubsection{Translation and Rotation Errors}

Table~\ref{tab:table1} tabulates the performance of the PoseNet with various front-end unimodel and multimodal feature extractors, along with the proposed hybrid learners. The results in this Table can be described in three subdivisions. The primary section from M$1$ to M$7$ is the outcomes of the unimodal PoseNet with unimodality-based feature extractors. The subsequent section extending from M$8$ to M$12$ depicts the performances of five multimodality-based learners. M$8$ represents late fusion (LF), M$9$ (AEF$^{ResNet101}$) and M$10$ (MEF$^{ResNet101}$) represent early fusion on ResNet101 as feature extractor with addition and multiplication, respectively. M$11$ (AEF$^{VGG19}$) and M$12$ (MEF$^{VGG19}$) are the results for early fusion on VGG$19$ with addition and multiplication, respectively. The third section consists of proposed hybrid learners, where M$13$, M$14$, and M$15$ stand for AHL that combines the early fusion models, M$9$ and M$11$, MHL, which combines the early fusion models, M$10$ and M$12$, and the HLFF obtained after averaging the predicted scores of the four models, M$9$, M$10$, M$11$, and M$12$. The results are computed as the mean and median values of translation and rotation errors. The translation errors are measured in terms of meters ($m$), while the rotation error is measured in degrees $(^\circ)$.
% The results achieved from this experiment are derived by applying the evaluation metrics as explained in the equations (\ref{eq:1}), (\ref{eq:2}), and (\ref{eq:3}). 
\begin{figure}[t]
     \centering
     
     \begin{subfigure}[t]{\columnwidth}
        \centering
         \includegraphics[trim={1.0cm, 0.5cm, 1.0cm, 1.2cm}, clip, width=\columnwidth]{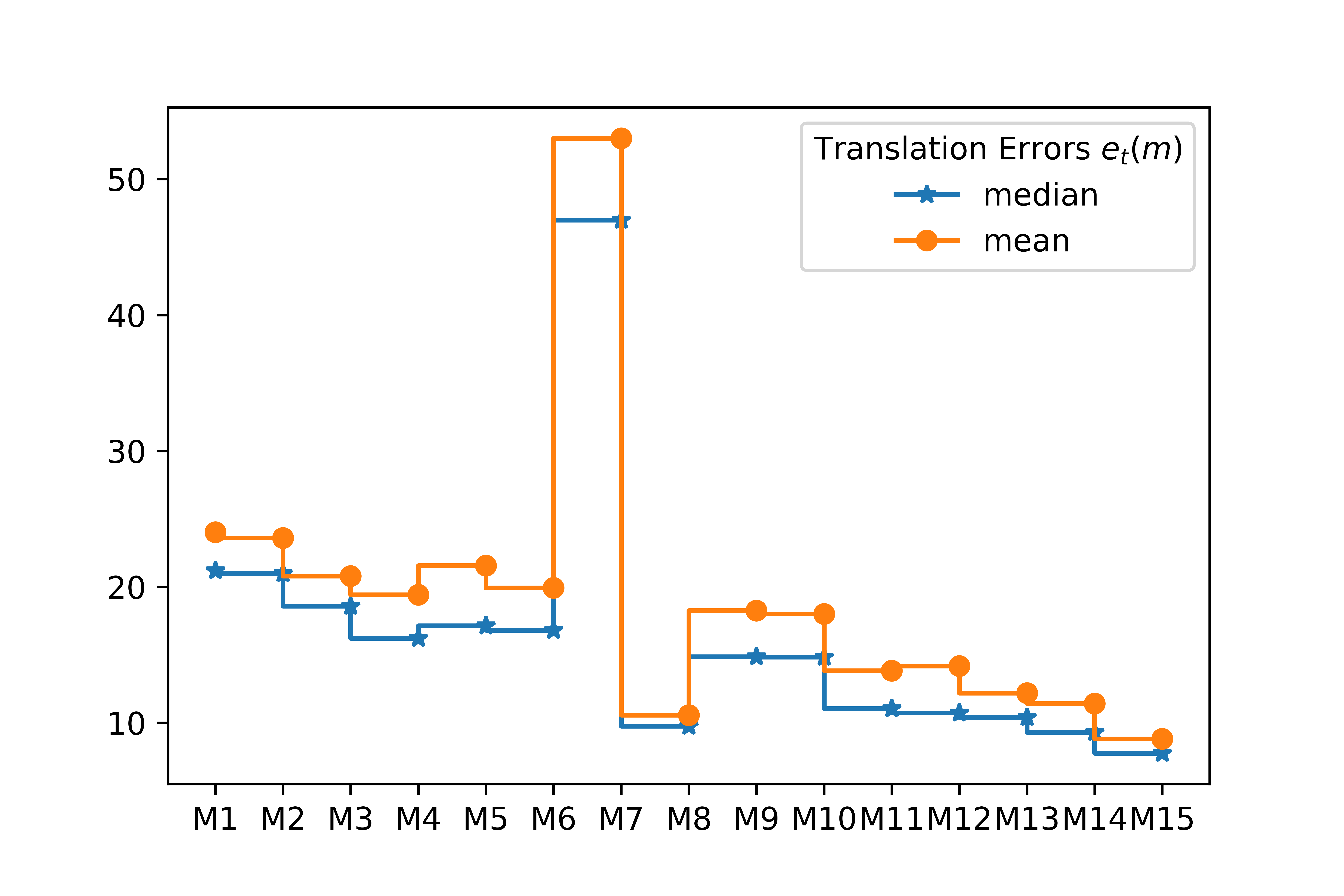}
             \caption{Error in Terms of Translation.}
             \label{fig:y equals x}
         \end{subfigure}
         \hfill
         %-----------------------------------
         \begin{subfigure}[t]{\columnwidth}
             \centering
              \includegraphics[trim={1.0cm, 0.5cm, 1.0cm, 1.0cm}, clip, width=\columnwidth]{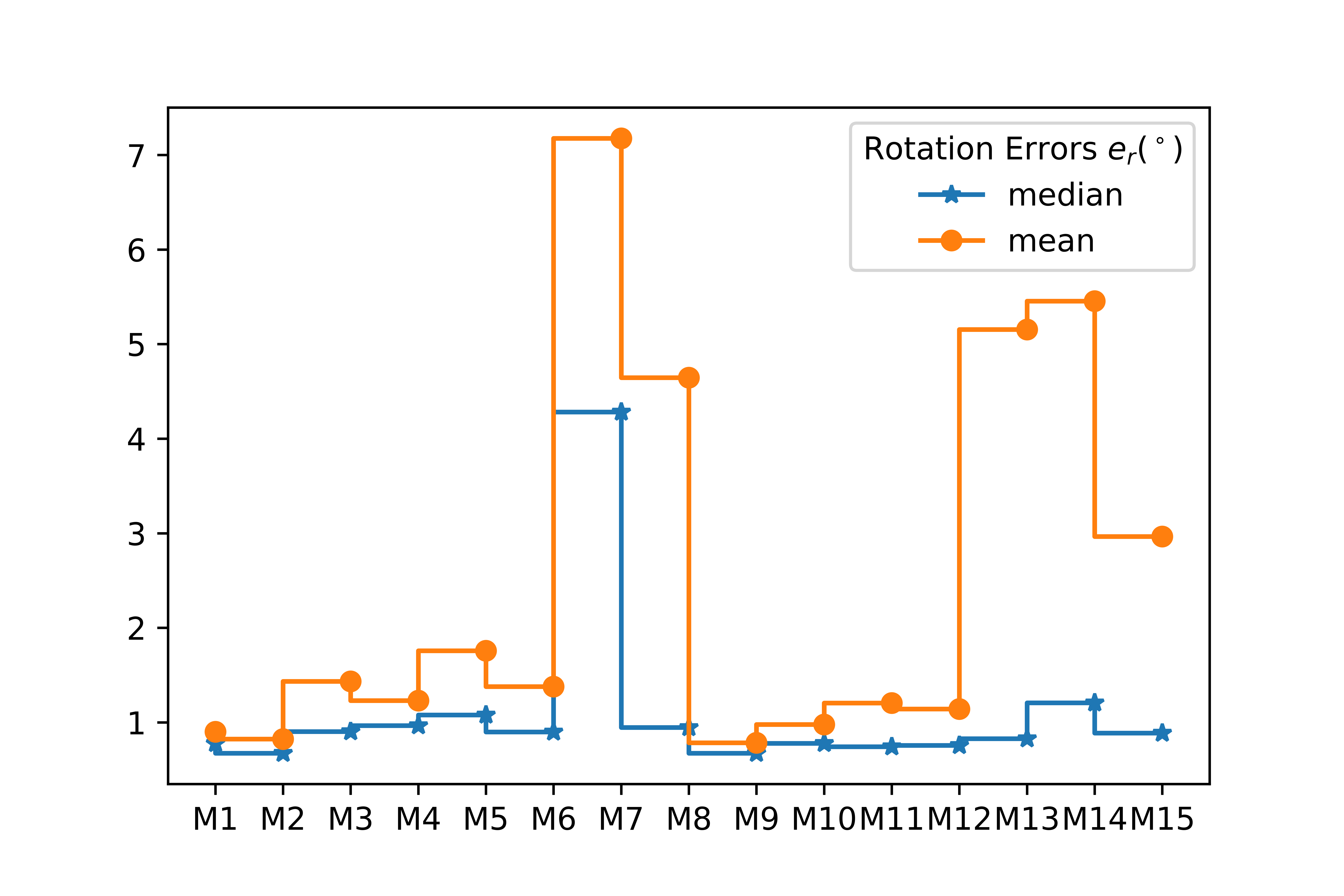}
             \caption{Error in Terms of Rotation.}
             \label{fig:five over x}
         \end{subfigure} 
         \hfill
         %---------------------------------
        \begin{subfigure}[t]{\columnwidth}
            \centering
              \includegraphics[trim={0.90cm, 0.5cm, 1.0cm, 1.0cm}, clip, width=\columnwidth]{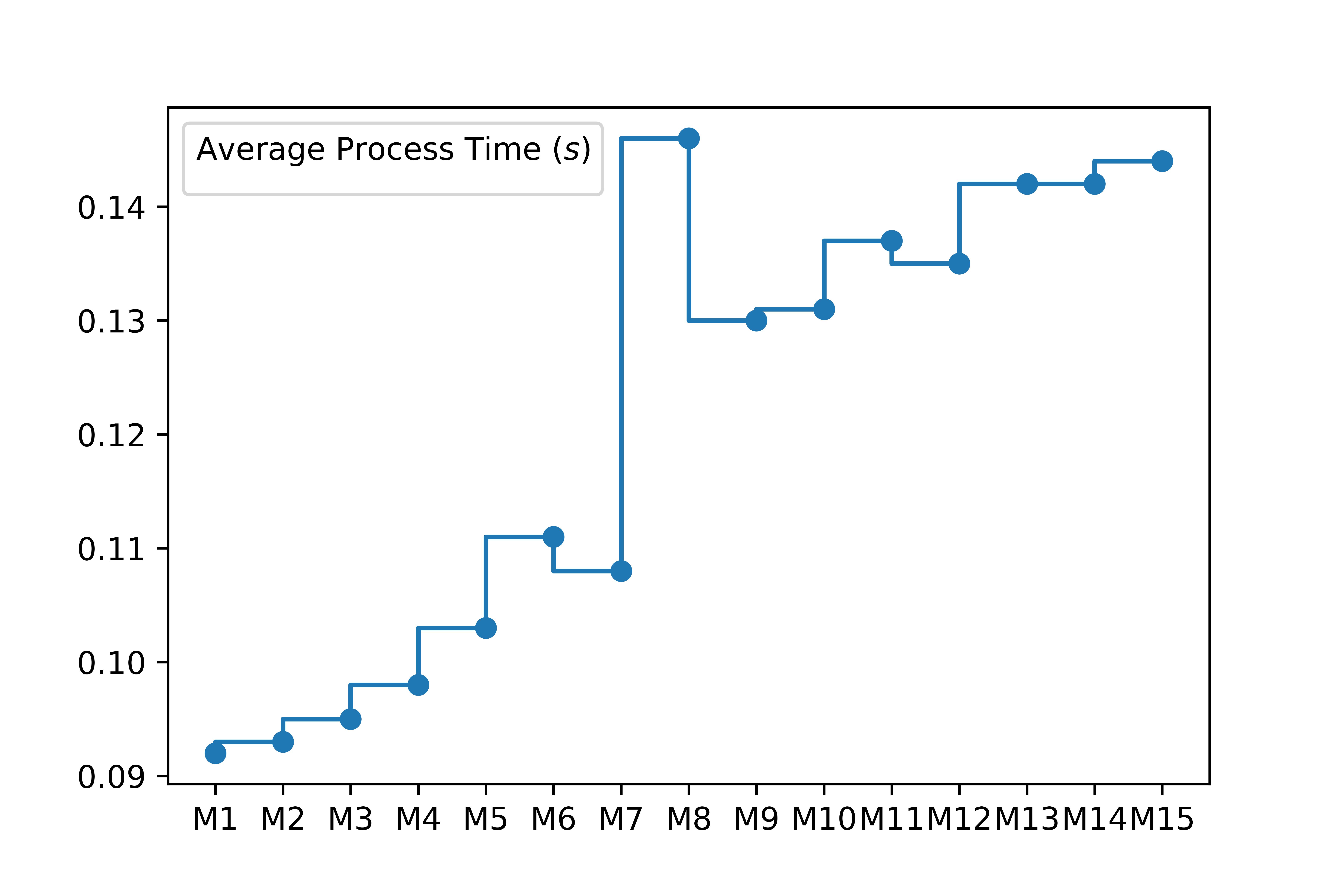} 
             \caption{Average Processing Time per Sample.}
             \label{fig:five over time}
         \end{subfigure}
         \hfill
            \caption{Performance Analysis of All Different Models.}
            \label{fig:performance}
        
\end{figure}    
Considering the unimodal-based PoseNet implementation, it is very apparent that ResNet$101$ and VGG$19$ give the best two outcomes among others. 
% While making the comparison we first compare the unimodal with multimodality learning. 
In terms of translation error, the late fusion model shows better performance than unimodality-based learners, but not good as compared to the early fusion model. Let's pick the ResNet$101$-based PoseNet as baseline model for rest of the comparative analysis because it has the best performance amongst the all the unimodals. Here, the late fusion shows a $66$\% decrease in the translation's median errors and an $84$\% decrease in the translation's mean errors when compared to ResNet$101$ (M$4$). The median of rotation errors in the late fusion model shows a decrease of $2$\% but the mean of rotation errors increase by $73$\%.
% when compared to our unimodality baseline model (M$4$). 

On comparison of the baseline model (M$4$) with the early fusion model using addition having ResNet101 as a feature extractor (M$9$), it is seen that there is a $9$\% decrease in translation's median error and a $6$\% decrease in translation's mean error. On the other hand, the median of rotation error shows a $43$\% decrease and mean of rotation error shows a $41$\% decrease. The comparison with the early fusion model (M$10$) using multiplication on VGG$19$ exhibits a $51$\% decrease in translation's median error and a $37$\% decrease in translation's mean error. While the rotation errors drop in median value by $28$\% and in the mean value by $8$\%.

It is evident from the Table~\ref{tab:table1} that hybrid learners show much better performance than the unimodal and early fusion models. The hybrid learner using average filtering shows a $109$\% decrease in the translation's median errors and a $120$\% decrease in the translation's mean error when compared to the ResNet$101$-based PoseNet. While for rotation errors, there is a $4$\% increase in the median and a $73$\% increase in the mean.

In holistic analysis, it is observed that the late fusion shows an improvement of $37$\% in translation and $3$\% in rotation. On considering the early fusion model using ResNet as a feature extractor, the translation shows $6\%$ improvement while rotation shows $31$\%. The improvement of the early fusion using VGG$19$  as a feature extractor in terms of translation and rotation is $31$\% and $24$\%, respectively. 
It is quite evident from the Table~\ref{tab:table$2$} that the proposed HL has a negligible low results for rotation with a decrease of $4$\% nevertheless, it is the best model considering a huge improvement in translation by $50$\% across all the modals. 

\begin{table}[t]
 \begin{center}
 \setlength\tabcolsep{8 pt}
 \begin{tabular}{|l|c|c|c|c|}
 \hline
 
\multirow{2}{*}{\textbf{Model Name}} & \multicolumn{2}{c|}{\makecell{\textbf{Improvement in}}} & \multirow{2}{*}{\makecell{\textbf{Timing} \\\textbf{Overhead} $(ms)$}} \\
\cline{2-3}
&  $e_t$ (\%) & $e_r$ (\%) & \\ 
\hline\hline
LF & $37$ & $3$ & $48$ \\
\hline
EF$^{ResNet}$ & $6$ & $31$ & $36$ \\
\hline
EF$^{VGG}$ & $31$ & $24$ & $39$ \\
\hline
HLFF & $50$ & $-4$ & $46$ \\
\hline \hline
\end{tabular}
\end{center}
\centering
\caption{Performance Improvement of Proposed Hybrid Learner When Compared to the Baseline PoseNet with ResNet101 as Front-end.}
%\vspace{-0.3cm}
\label{tab:table$2$}
\end{table}

\subsection{Timing Analysis}

% \pgfplotstableread{
% thread  training validation
% 1   1.163   0.917
% 2   1.255   0.930
% 3   1.248   0.914
% 4   1.334   0.926
% 5   1.277   0.975
% 6   1.286   0.955
% 7   1.142   0.874
% 8   5.217   4.212
% 9   1.931   1.327
% 10  1.955   1.382
% }\datafile

% \begin{figure}[!ht]
% \begin{tikzpicture}
% \begin{axis}
% [
%     ylabel=Time (s),
%     xtick = \empty,
%     grid style = dashed,
%     extra x ticks = {1,2,3,4,5,6,7,8,9,10},
%     extra x tick labels = {M1,M2,M3,M4,M5,M6,M7,M8,M9,M10},
%     extra x tick style={
%           tick label style={rotate=0}, anchor=},
%     legend style={at={(0.09,0.8)},anchor=west}]
    
% \addplot table[x=thread,y=training] {\datafile};
% \addlegendentry{Training};
% \addplot table[x=thread,y=validation] {\datafile};
% \addlegendentry{Validation};
% \end{axis}
% \end{tikzpicture}
% \centering\caption{Timing Analysis.}
% % \vspace{-0.5cm}
% \label{fig:timing}
% \end{figure}

The timing analysis is conducted on a machine that uses an Intel Core i5 processor that uses the Google Colaboratory having a GPU chip Tesla K80 with 2496 CUDA cores, a hard disk space of 319GB and 12.6GB RAM. Table~\ref{tab:table1} shows the mean average processing time calculated for processing a batch of ten samples.

As seen from the Table~\ref{tab:table1} and Figure~\ref{fig:performance}, the fusion models which involve early, late, and hybrid learner take slightly extra time compared to the unimodality-based baseline PoseNet. The late fusion model (M8) takes more processing time in comparison to all the other models, as it uses five pretrained modalities, which are trained and tested individually, thereby, increasing the time overhead. The early fusion models also show an increase in the processing time in comparison to the unimodels but lesser than the late fusion model, as training the pretrained model after weight enhancement takes more time. The hybrid learners also show the same trend because of the underlying fact that it is a combination of the early and late fusion methods. These models employ weight enhanced early fusion models adding to the time overhead, besides fusing the scores from different models after validation.

Note that the hyper-parameters have been fixed throughout the experimental analysis on various models to avoid uncertainties in the comparative study. The learning rate used in all the models is $0.01$, dropout rate for the dropout layer of the PoseNet is set to $0.5$, and the batch size during training is fixed to $34$. Hence, every model is trained for $1000$ epochs with Adam optimizer.
% to show the consistent comparison across distinctive models.

\section{Conclusion}\label{conclusion}

This work introduces a the hybrid learner to improve the localization accuracy of a pose regressor model for SLAM. 
% The unimodal and multimodality-based learner forms the bases for the hybrid learner. Unimodal is a simple approach utilizing seven distinctive DCNNs. The multimodality-based learner uses two techniques, which are early fusion and late fusion. Multimodality-based learner provides more favorable performance than unimodal learners. 
The hybrid learner is a combination of multimodal early and late fusion algorithms to harness the best properties of the both. The extensive experiments on the Apolloscape self-localization dataset show that the proposed hybrid leaner is capable of reducing the translation error nearly by a $50$\% decrease, although the rotation error gets worse by a negligible $4$\% when compared to unimodal PoseNet with ResNet$101$ as a feature extractor. 
% This increase in the rotation values accounts for one of the significant shortcomings of our experimental analysis. The experiments gradually improved the translation errors but, at the same time, increased rotation errors. While considering a holistic approach, it is evident that the hybrid model using averaging is the best performing model as the improvement in the accuracy is drastic. 

Thus, the future work aims at minimizing the rotation errors
% as a significant improvement in the translation is already achieved, thereby working towards the shortcomings of the project. 
and overcoming the little overhead in the processing time. 
% It is visible from the timing analysis that the models involving the multimodal-based learner and hybrid learner consume more time than all unimodal-based learners, thereby making the best performing model less efficient in time analysis. 

% performed on the compared models. From the results in Table~\ref{tab:table1}, ResNet101 and VGG19 are the best two pretrained models that give better accuracy when compared to the other individual pretrained models. 

% The results obtained in Table~\ref{tab:table1}, also prove that score fusion shows better results for translation but poor results for rotation error. The training time weight enhancement model for addition and multiplication give better rotation error when compared to score fusion. While, the results for translation error of training time fusion are not better than the score fusion results, they are good when compared to the results obtained for the individual models. From the performance results of all the models, it can be concluded that fusion of different models give better accuracy than just using one model for predictions. The future work is dedicated to overcome the additional processing time used for the fusion operations. 

% We plan on performing more operations for fusion of models including min-max, concatenation and PCA optimization to determine the best method that can be used for the fusion of SLAM. The concept of transfer learning using various algorithms for fusion are still to be expanded.

\section*{Acknowledgment}

This work acknowledges the Google for generosity of providing the HPC on the Colab machine learning platform and the organizer of Apollo Scape dataset.

\bibliography{citation}
\bibliographystyle{ieeetr}

\end{document}